\documentclass[runningheads]{llncs}
\usepackage{graphicx}

\usepackage{tikz}
\usepackage{comment}
\usepackage{amsmath,amssymb} 
\usepackage{color}
\usepackage[colorlinks,linkcolor=red]{hyperref}

\begin{document}
\pagestyle{headings}
\mainmatter
\def\ECCVSubNumber{100}  

\title{Bias Elimination for Domain Adaptive Pedestrian Re-identification} 

\titlerunning{Bias Elimination for Domain Adaptive Pedestrian Re-identification}
%
\author{Jianyang Gu\inst{1,2} \and
Hao Luo\inst{2} \and
Weihua Chen\inst{2} \and
Yiqi Jiang\inst{2} \and
Yuqi Zhang\inst{2} \and \\
Shuting He \inst{1,2} \and
Fan Wang\inst{2} \and
Hao Li\inst{2} \and
Wei Jiang\inst{1}}
\authorrunning{J. Gu et al.}
%
\institute{Zhejiang University \and
Alibaba Group\\
\email{gu\_jianyang@zju.edu.cn}\\
\email{\{michuan.lh,kugang.cwh\}@alibaba-inc.com}}
\maketitle

\begin{abstract}
This paper presents our proposed methods for domain adaptive pedestrian re-identification (Re-ID) task in \href{http://ai.bu.edu/visda-2020/}{Visual Domain Adaptation Challenge (VisDA-2020)}. 
Considering the large gap between the source domain and target domain, we focused on solving two biases that influenced the performance on domain adaptive pedestrian Re-ID and proposed a two-stage training procedure. 
At the first stage, a baseline model is trained with images transferred from source domain to target domain and from single camera to multiple camera styles. 
Then we introduced a domain adaptation framework to train the model on source data and target data simultaneously. 
Different pseudo label generation strategies are adopted to continuously improve the discriminative ability of the model. 
Finally, with multiple models ensembled and additional post processing approaches adopted, our methods achieve 76.56\% mAP and 84.25\% rank-1 on the test set. 
Codes are available at \href{https://github.com/vimar-gu/Bias-Eliminate-DA-ReID}{https://github.com/vimar-gu/Bias-Eliminate-DA-ReID}. 
\end{abstract}

\section{Introduction}

Pedestrian re-identification (Re-ID) aims to match specific person identities across multiple cameras. 
As more and more surveillance cameras are being deployed in cities, pedestrian Re-ID can play an indispensable role in modern security systems. In recent years, deep learning methods have made a significant progress on pedestrian Re-ID task \cite{Luo_2019_Strong_TMM,Luo_2019_CVPR_Workshops,sun2018beyond}. 
However, pedestrian Re-ID still faces many challenges, one of which is the large data amount. 
As the pedestrian Re-ID task is an open-set problem, it is impossible to manually label all the pedestrian images produced by surveillance cameras day by day. 
Based on the situation, domain adaptation has attracted much attention on the pedestrian Re-ID task \cite{deng2018image,zhang2019self,fu2019self,chen2019instance,song2020unsupervised}. 

Compared with traditional domain adaptation tasks, domain adaptive pedestrian Re-ID is much harder, as the source domain and target domain share no identical classes. 
Moreover, in VisDA-2020, a synthetic dataset are provided as the labeled source domain \cite{sun2019dissecting} and a real-world dataset is adopted as the target domain, where exists a large domain gap. 
In this work, we analyzed the biases in domain adaptive pedestrian Re-ID task introduced by different datasets and different cameras, and proposed a domain adaptation framework to solve the problem. 

The rest of the paper is organized as follows. In Section \ref{methods}, the proposed methods is introduced. The experimental results are presented in Section \ref{experiments}. And finally Section \ref{conclusion} concludes the paper. 

\section{Methods}
\label{methods}
\subsection{Data Generation}

Domain adaptive pedestrian Re-ID task is faced with two main biases which will introduce disturbance to the discriminative ability of the model. 

The first one is the inter-domain gap between different datasets. 
In the challenge, the source domain contains synthetic pedestrian images while the target domain is consisted of realistic ones. 
The huge appearance difference between the two domains brings poor performance to directly using the model trained on source domain for testing. 
To bridge this domain gap, generative adversarial networks (GAN) is commonly adopted to transfer source domain images to target domain. 
In the challenge, an SPGAN-transferred dataset is provided \cite{deng2018image}. 
Through transferring, realistic texture from the target domain is added into the labeled source synthetic images. 
Therefore, conducting supervised learning on the SPGAN-transferred dataset can gain better discriminative ability on the target domain. 

The second one is the intra-domain bias introduced by different cameras, which indicate differences on orientation, illumination, occlusion, resolution and many more conditions. 
In this work, we introduced starGAN to produce images with different camera styles \cite{zhong2018camera,zhong2018generalizing}. 
With a large amount of additional images, the model can generalize better among images captured with different cameras. 
This part of data will be named CamStyle data in the following sections. 

\subsection{Baseline Model}
As the source domain provides labeled images while the target domain doesn't, we train the baseline model in supervised manner on source domain. 
Both SPGAN-transferred images and CamStyle images are utilized in the training process. 
Label-smoothed cross entropy loss is adopted for classification and soft-margin triplet loss is adopted for better clustering performance. 

\subsection{Domain Adaptation}
\begin{figure}
\centering
\includegraphics[width=10cm]{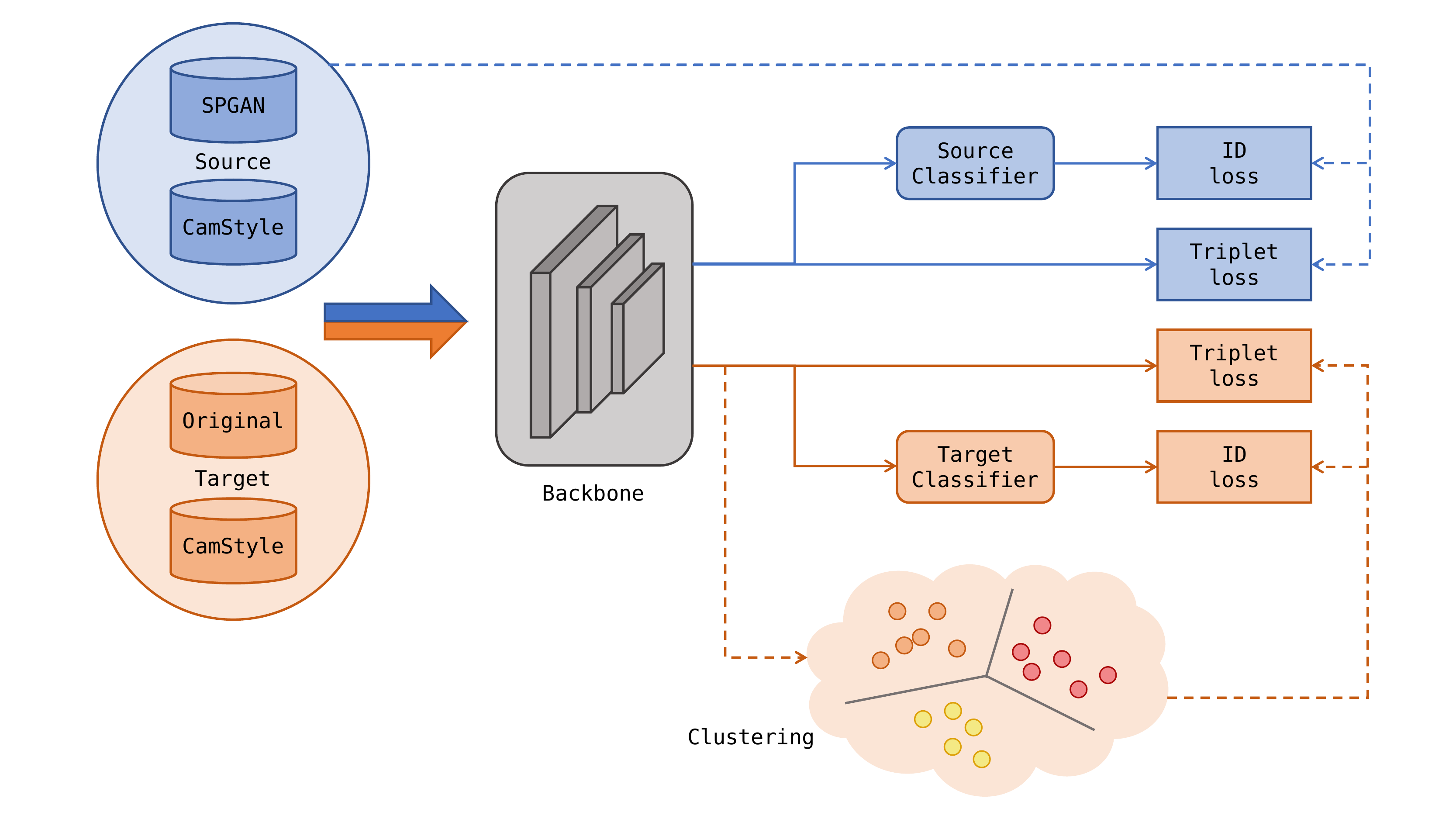}
\caption{The pipeline for domain adaptation stage }
\label{fig:da}
\end{figure}
To better eliminate the inter-domain gap in domain adaptation tasks, we introduced a domain adaptation framework to train the model simultaneously on source domain and target domain. 
We designed a model with backbone and different classifiers for each domain.
With this structure, the network can fully utilize the extracted feature to identify classes from each domain and narrow the feature distribution gap between these two domains. 
For the source domain, the training procedure is the same as that at baseline model training stage. 
And for the target domain, an additional clustering operation will be executed to produce pseudo labels. 
As the exact class number of target domain is unknown, DBSCAN is adopted as the clustering algorithm. 
The model pipeline is shown in Figure \ref{fig:da}, where blue lines indicate source domain data and orange lines indicate target domain data. 
The dashed lines represent the labels flow. 
For source domain, the labels come from the dataset, while for target domain, labels come from clustering. 
The embedded features extracted by backbone network are then passed through corresponding classifiers to get their classification scores. 

The domain adaptation training process is separated into two stages with different pseudo label generation strategies. 
At the first stage, we selected $500$ classes with most samples clustered as the target training set and discard the rest. 
And as the model can better discriminate different identities, the outliers are regarded as classes with few samples. 
So at the second stage, we added another $200$ classes each with one sample into the target training dataset. 
For the triplet loss calculating, these classes will only contribute to the loss of negative samples. 
Through the adoption of the two-stage pseudo label generation strategy, the model can continuously improve its performance. 

The clustering process is executed every $6$ epochs. 
After the pseudo label generation, the source domain data and target domain data are sampled at a certain rate each to form a mini-batch. 
And for each mini-batch, original data and CamStyle data are also sampled at a fixed proportion. 
That means a mini-batch is either composed of source domain data and target domain data, while contains both original data and Camstyle data. 

\subsection{Post processing}
After the features are extracted for testing set, we adopted several post-processing methods to further improve the model performance. 
The main focus of treatment is on the camera bias, which will largely influence the discriminative ability of the model. 
Firstly, the mean value of features under the same camera is calculated and subtracted from each feature. 
Then for each sample, the feature is updated with its closest neighbors. 
Considering there is no camera label provided in the testing set, we trained an additional camera model to predict the camera label for each image. 
Besides, inspired by \cite{voc_reid}, the features extracted by the camera model is utilized to calculate a camera distance matrix, which will be subtracted from the original feature distance matrix at a certain rate. 
Additionally, we built up a topology map representing the probability of showing up under a certain camera based on the given camera labels in validation set. 
Images under cameras with larger probability will be assigned larger distance weights. 
Traditional re-ranking \cite{zhong2017re} is also adopted to update the distance matrix. 

\section{Experimental Results}
\label{experiments}
\subsection{Implementation Details}
The model structure is based on \cite{Luo_2019_Strong_TMM,Luo_2019_CVPR_Workshops}. 
We added an additional linear layer after the feature is extracted to compress the feature dimension to $512$. 
At the baseline model training stage, a 700-class classifier conducts the classification operation for source domain, while at the domain adaptation stage, we designed two classifiers with corresponding dimensions to source domain and target domain. 
We trained the models on different backbones pretrained on ImageNet \cite{deng2009imagenet}, among which ResNet50-ibn-a \cite{pan2018two}, ResNet50-ibn-b \cite{pan2018two}, ResNet101-ibn-a \cite{pan2018two} and HRNetv2-w18 \cite{wang2020deep} showed better performance when testing on target domain. 
For data augmentation, we used random horizontal flip, padding and erasing \cite{zhong2020random}. 

We utilized SGD optimizer with a original $0.02$ learning rate. Warm-up strategy is adopted during the first $10$ epochs, and the learning rate is decayed at the $24th$ and $48th$ epochs. The model is trained for $60$ epochs in total. 

For the standard model training, the images are resized to $384\times 128$. 
We also finetuned multiple models with different image sizes based on standard models to compare features in more scales. 
Specifically, $384\times 192$ is selected by us for larger image size.

And for camera models, we utilized ResNet101 \cite{he2016deep}, ResNet152 \cite{he2016deep}, ResNet101-ibn-a \cite{pan2018two} and HRNetv2-w18 \cite{wang2020deep}. 
The camera distance matrix used in post processing step is generated from the mean of all camera distance matrixes. 
Finally we calculated a weighted sum of Re-ID distance matrixes and conducted the post processing methods to achieve the final score. 
Further details can be found at \href{https://github.com/vimar-gu/Bias-Eliminate-DA-ReID/blob/master/VisDA.md}{Reproduce Instructions}. 

\subsection{Ablation Study}
\subsubsection{Effectiveness of Individual Components}

\setlength{\tabcolsep}{4pt}
\begin{table}
\begin{center}
\caption{The model performance on validation set with different components. Note: by our experience, there can be a large fluctuation of validation scores which are not completely positive correlated to the scores on testing set }
\label{table:stages}
\begin{tabular}{lllll}
\hline\noalign{\smallskip}
Methods & mAP & Rank-1 & Rank-5 & Rank-10\\
\noalign{\smallskip}
\hline
\noalign{\smallskip}
Direct transfer & 16.2 & 32.4 & 50.7 & 58.6 \\
+ SPGAN & 24.7 & 44.6 & 62.3 & 72.4 \\
+ CamStyle & 30.7 & 59.7 & 77.5 & 83.3 \\
+ Domain Adaptation & 44.9 & 75.3 & 86.7 & 91.0 \\ 
+ Finetuning & 48.6 & 79.8 & 88.3 & 91.5 \\ 
+ Post Processing & 70.9 & 86.5 & 92.8 & 94.4 \\
\hline
\end{tabular}
\end{center}
\end{table}

We compare the evaluation result on validation set to demonstrate the effectiveness of each component in our model structure and the experimental results are summarized in Table \ref{table:stages}. In the table, "Direct Transfer" means testing on target domain validation set with model trained on original source domain. It can be seen that directly applying the model trained on source domain to target domain shows poor performance, with a 16.2\% mAP and 32.4\% Rank-1. Through the introduction of SPGAN-generated data, part of the domain gap has been narrowed. And by adding extra CamStyle data, the performance is boosted by large margin. It shows that although diminishing the domain gap is effective, the camera bias is also an important issue. 

The domain adaptation process further reduces the inter-domain gap, with about 15\% growth on mAP and Rank-1. And during the finetuning stage with more samples utilized, there is an additional 4\% increase. The post processing methods also play a significant role in the model performance. The experiment is executed on ResNet50-ibn-a backbone, and the stats are similar on the other backbones. Note that the stats are obtained offline, so there might be some differences under different evaluation systems. 

\begin{figure}
\centering
\includegraphics[width=10cm]{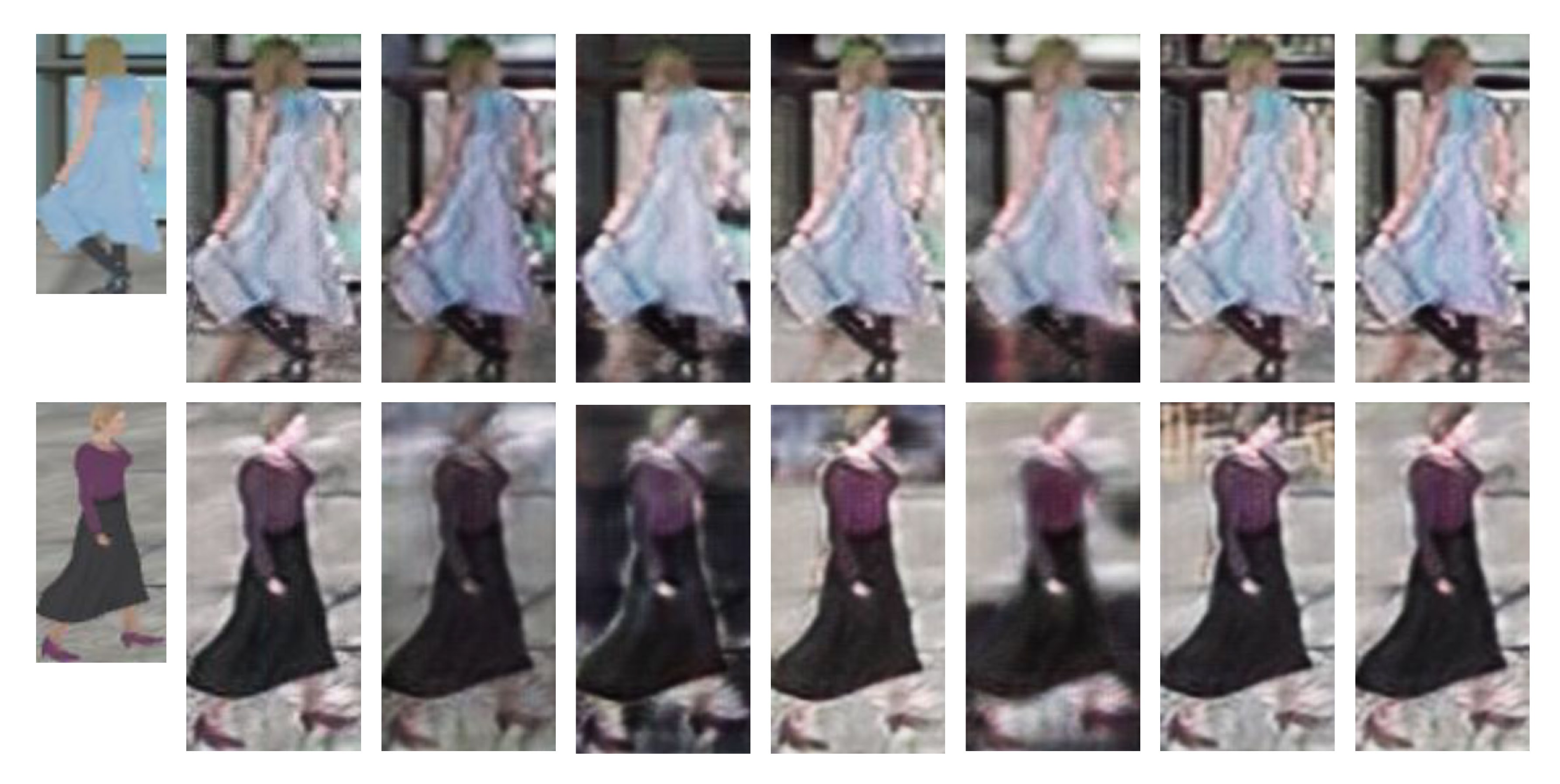}
\caption{The generated training data. The first column is from source domain. The second column is from SPGAN-generated data and the other columns are from CamStyle data. }
\label{fig:camstyle}
\end{figure}

\subsubsection{Visualization of Generated Data}
We selected some of our generated data utilized in training processes to show the influence from data more intuitively. 
From Figure \ref{fig:camstyle}, it can be observed that through the introduction of SPGAN, a large number of texture is added into the images. 
Add the CamStyle data further improves the diversity of data, where a variety of resolution, illumination conditions are simulated. 

\subsubsection{Effectiveness of model ensemble}

\setlength{\tabcolsep}{4pt}
\begin{table}
\begin{center}
\caption{The performance on testing set with single model and model ensemble}
\label{table:ensemble}
\begin{tabular}{lllll}
\hline\noalign{\smallskip}
Methods & mAP & Rank-1 & Rank-5 & Rank-10\\
\noalign{\smallskip}
\hline
\noalign{\smallskip}
Single Model & 71.1 & 79.8 & 86.8 & 90.4 \\
Model Ensemble & 76.6 & 84.3 & 89.6 & 92.4 \\
\hline
\end{tabular}
\end{center}
\end{table}
With the above generated data and training stages, our best model (ResNet50-ibn-a) can reach about 71.1 mAP and 79.8 Rank-1. We also trained the model with ResNet50-ibn-b, ResNet101-ibn-a and HRNetv2-w18 backbones with different image sizes. Finally we integrated all models and adjusted some post-processing parameters to gain more than 5.5\% mAP and 4.7\% boost compared to the mean performance of all backbones. Note that our final Rank-1 is the highest among the competitors. 

\section{Conclusion}
\label{conclusion}
We have presented our framework for the domain adaptive pedestrian Re-ID challenge. It mainly focuses on the domain gap and camera bias which would influence the discriminative ability of models in the task. The top performance during the challenge had proved the effectiveness of our proposed methods, which can be further analyzed and better utilized in future works. 

%
%
\bibliographystyle{splncs04}
\bibliography{egbib}
\end{document}